# "*A net for everyone*": fully personalized and unsupervised neural networks trained with longitudinal data from a single patient

Christian Strack[1,2*], Kelsey L. Pomykala[3], Heinz-Peter Schlemmer[1,5], Jan Egger[3,4]
Jens Kleesiek[3,4,5]

[1] Division of Radiology, German Cancer Research Center (DKFZ), 69120 Heidelberg, Germany

[2] Medical Faculty Heidelberg, Heidelberg University, 69120 Heidelberg, Germany

[3] Institute for AI in Medicine (IKIM), University Hospital Essen (AöR), Girardetstraße 2, 45131 Essen, Germany

[4] Cancer Research Center Cologne Essen (CCCE), University Medicine Essen, Hufelandstraße 55, 45147 Essen, Germany

[5] German Cancer Consortium (DKTK), Partner Site Essen, Hufelandstraße 55, 45147 Essen, Germany

*corresponding author: c.strack@dkfz-heidelberg.de

## Abstract

With the rise in importance of personalized medicine, we trained personalized neural networks to detect tumor progression in longitudinal datasets. The model was evaluated on two datasets with a total of 64 scans from 32 patients diagnosed with glioblastoma multiforme (GBM). Contrast-enhanced T1w sequences of brain magnetic resonance imaging (MRI) images were used in this study. For each patient, we trained their own neural network using just two images from different timepoints. Our approach uses a Wasserstein-GAN (generative adversarial network), an unsupervised network architecture, to map the differences between the two images. Using this map, the change in tumor volume can be evaluated. Due to the combination of data augmentation and the network architecture, co-registration of the two images is not needed. Furthermore, we do not rely on any additional training data, (manual) annotations or pre-training neural networks. The model received an AUC-score of 0.87 for tumor change. We also introduced a modified RANO criteria, for which an accuracy of 66% can be achieved. We show that using data from just one patient can be used to train deep neural networks to monitor tumor change.

*Keywords*: Neural networks, personalized, Wasserstein-GAN, unsupervised, machine learning, privacy-safe, zero-training data, longitudinal, brain tumor, MRI.

# 1      Introduction

One key difference between human and artificial intelligence is the number of training examples needed to generate knowledge. While children can learn to recognize new objects with only a few examples, most machine learning tasks require hundreds of examples for the same task. In fact, increasing the dataset size is often a key step in improving the performance of a machine learning model. ImageNet [1], the most famous dataset in computer vision, now consists of over 14 million training examples. The state of the art models in computer vision are often trained on large datasets such as ImageNet and may not transfer well to smaller datasets. Getting large datasets may not always be a feasible approach though, especially in the medical domain.

Gathering large datasets is one of the key challenges of medical deep learning applications. Keeping a patient's medical information safe is critical and there are laws protecting it in most countries. This makes it more difficult to get the data and leads to the medical datasets being much smaller compared to traditional computer vision datasets. Additionally, deep neural networks themselves offer another privacy threat. It has been shown that training examples of fully trained networks can be recovered with a model inversion attack [2]. This makes it more difficult to publish medical deep learning applications as the patient's privacy can not be guaranteed. These two reasons give a big incentive to find ways to train neural networks with smaller datasets or even just one patient's data.

There have been several models proposed to challenge the task of reducing the number of training examples. One shot learning is a method of learning a class from only one labeled example [3]. Siamese neural networks are able to determine if two images show the same person, even if they have never seen images of that person before [4]. They have also been used in medicine to distinguish between COPD and asthma [5]. While new classes can be learned from as little as one example, one shot learning still requires thousands of training examples of other classes beforehand. Another method to handle small datasets is transfer learning, where networks trained on large datasets are used as a starting point to train on training examples of new classes. Transfer learning makes use of the fact that features learned on the large dataset can be reapplied to new data.

In this paper, we introduce personalized neural networks, which use only one patient's data for training. Our proposed method only needs two MRIs from the same patient and no additional pretraining. This also results in a privacy-safe processing of the data, because the data "stays" within the same patient. Our model is based on Generative Adversarial Networks (GANs) [6]. GANs have gained in popularity in recent years in the medical AI community. Originally used for image synthesis, there have been applications to generate medical images [7, 8]. Other studies focus on classification or segmentation tasks [9, 10]. We apply the personalized neural networks on subjects with brain tumors.

Brain tumors belong to the most devastating diagnoses, in particular for a confirmed glioblastoma multiforme (GBM) [11]. Despite massive research efforts and advancements in other cancer types, like breast cancer [12] or prostate cancer [13], the life expectancy of a confirmed GBM with treatment, including chemotherapy, radiotherapy and surgery, is still only around one year [14].

Nevertheless, disease progression and treatment decisions are strongly dependent on maximum tumor diameter and tumor volume, as well as the corresponding morphological changes during a treatment period. The imaging method of choice here is magnetic resonance imaging (MRI). However, MRI does not provide any semantic information for brain structures or the brain tumor per se. This has to be done manually, semi-manually or automatically, in a post-processing step, commonly referred to as a *segmentation*. Manually performed, however, a segmentation is very time-consuming and operator-dependent, especially when performed in a three-dimensional image volume [15], which needs slice-by-slice contouring. Hence, an automatic (algorithmic) segmentation is desired, especially when large quantities of data volumes have to be processed. Even if it is still considered an unsolved problem, there has been steady progress from year to year; and data-driven approaches, like deep neural networks, currently provide the best (fully automatic) results. However, a segmentation with a data-driven approach, like deep learning [16], comes with several burdens: Firstly, the algorithm generally needs massive annotated training data. Additionally, for inter-patient disease monitoring, several segmentations have to be performed, and in addition, these scans have to be registered to each other (which also adds uncertainty to the overall procedure, especially when deformed soft-tissue comes into play [17]). In this regard, we want to tackle these problems with a personalized neural network that needs just the patient's data, no annotations and no extra registration step. To the best of our knowledge, this is the first study using this little training data to train a deep neural network in the medical domain. The method addresses the issues of gathering big datasets in medicine and producing a privacy-safe network. The approach is considered as unsupervised learning as no data annotation is necessary. We evaluate the model with an ROC analysis as well as modified RANO criteria on two different datasets of longitudinal MRI images of patients with glioblastoma.

## 2 Methods

### 2.1 Model architecture and training

The neural network architecture used in this study is based upon Wasserstein GANs [18]. This is a modified version of Generative Adversarial Networks (GAN) [6]. These are a form of deep neural networks in which two sub-models are trained adversarily in a sum-zero game. A generator is trained to create new images, while a discriminator is trained to distinguish between real and synthetic images. In Wasserstein GANs the discriminator is modified to a critic function which leads to more stable training [18].

Our network architecture is similar to the model used by Baumgartner et al [19]. The aim of the network is to create a map which transforms an image from the first timepoint (t1) to the second timepoint (t2). This will make the model learn to represent the changes between the images, more specifically tumor growth/reduction in our case. To do this, augmented versions of the image at t1 are used as input to the generator. The generator will try to create a map that, when added to the input image creates an image of t2. The critic will try to distinguish these generated synthetic t2 images from the real t2 images. Thereby forcing the generator to learn the differences between the two timepoints.

The generator is based on the U-Net [20] structure. The U-Net is a fully convolutional network consisting of a contracting path (encoder) and an expanding path (decoder) with skip connections at each resolution level. It produces an output image of the same size as the input image, which is important in this case. The network structure is shown in more detail in Figure 1. A random slice of the third dimension was taken during each training step, so that the network received an input size of 256x256 pixels. For the ultimate prediction after training, the result for each of the 128 slices was calculated, saved and concatenated to the final 256x256x128 pixels volume. The critic function is also a fully convolutional network. Like in Baumgartner et al. [19], we used an architecture similar to the C3D network [21]. This is an encoder type architecture which produces a single value output (Figure 1 in the Supplement).

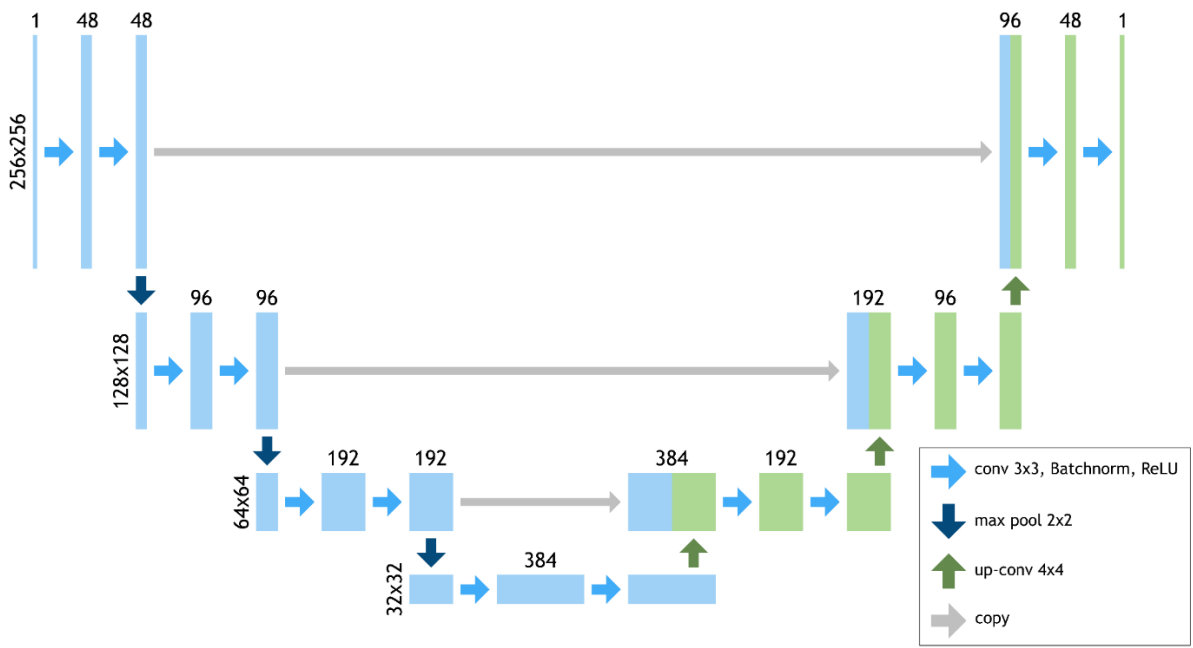

Figure 1: Architecture of the generator network. A U-Net structure is used. At each level there are two blocks of 3x3 convolution, Batchnorm and ReLU. 2x2 Max pool functions are used for downsizing. 4x4 transposed convolutions with stride=2 are used for upsizing. The size of the image at each level is shown on the left. The number of features in each block is shown on the top of the block.

The network was trained for 1000 epochs. In every epoch we updated the critic five times before updating the generator. In the first 25 epochs and every 100 epochs, the critic was updated 100 times. We used gradient penalty and the ADAM optimizer during training [22, 23]. Figure 2 gives an overview over the whole training process.

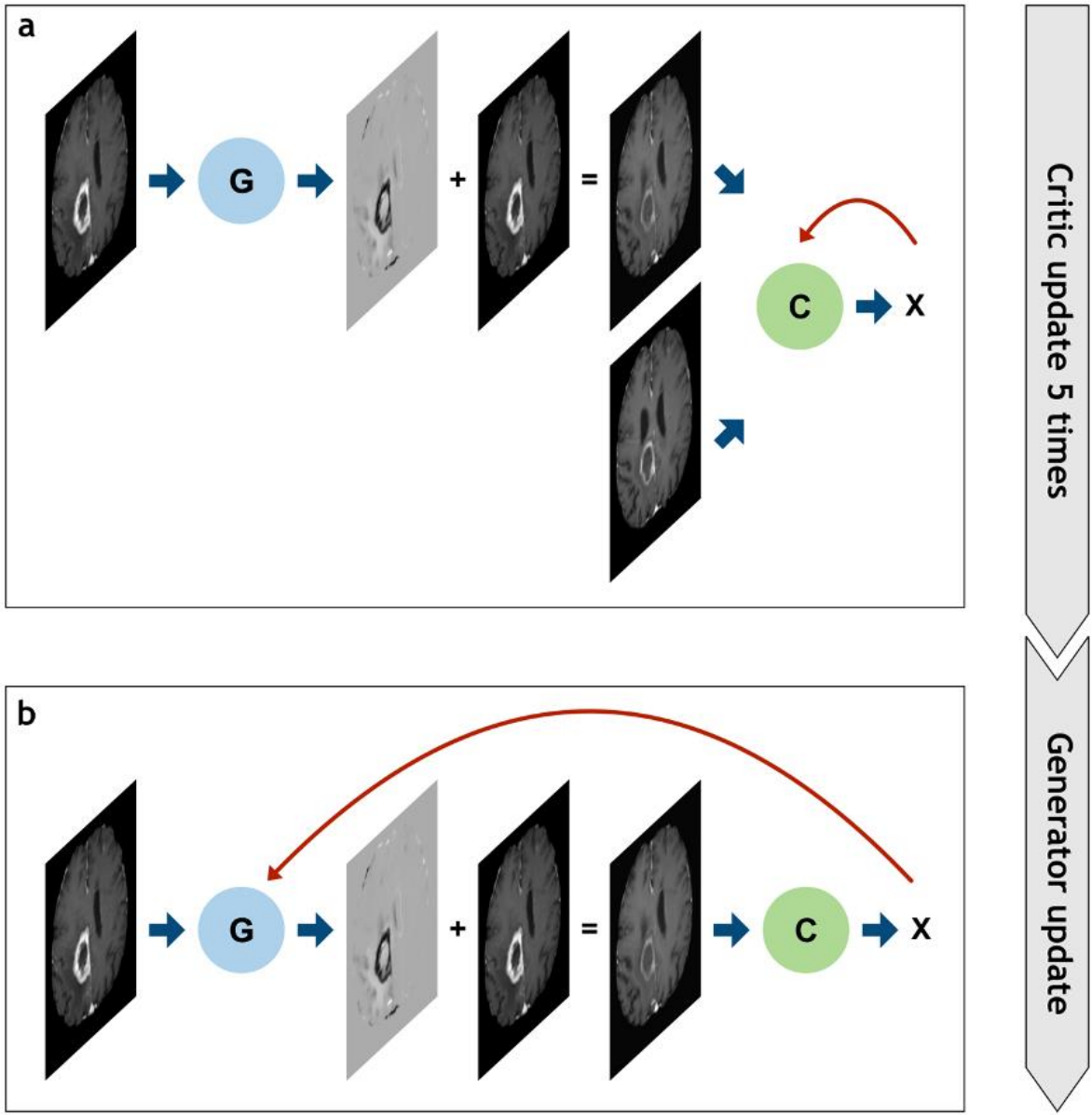

Figure 2: Overview over one training epoch. In (a) the critic function is trained. A t1 image is passed through the generator. The generator's output is a map which gets added to the t1 image. This produces the fake t2 image. The real and the fake t2 images are then passed to the critic. The output of the critic is incorporated into a loss function and backpropagated to update the weights of the critic network. In (b) the generator is trained. Again, a t1 image is passed to the generator. The output is added to the t1 image to create the fake t2 image. This is passed to the critic. The output is incorporated into the generator loss function and backpropagated through both networks to update the generator network.

During training, we discovered that the training process could be unstable when the two images were too similar or even identical. Therefore, we added a small square of 10x10 pixels of noise to a fixed position in one of the images. The noise was created by transforming Gaussian noise with a 2D Gaussian filter. The position of the noise was changed twice during the training (after 40% and 60% of all the training epochs). The concrete positions were at 50%, 35% and 65% of the size of the input image in both dimensions. Finally, after 80% of the epochs, the noise was removed completely for the rest of the training. After each change of the position of the noise, a model was saved. After the training finished, we took an ensemble of all the models, averaging over the results, disregarding those pixels that had been artificially changed in that part of the training.

## 2.2 Preprocessing

There were several preprocessing steps in this study. First, all images were resampled to 256x256x128 pixels. In MRIs, the pixel values obtained differ for identical tissues when different scanners are used. To deal with this problem, we histogram matched the images to each other. This was done using the histogram matching tool of 3D Slicer [24]. Next, the images were normalized to a range between 0 and 1. The brain of the patient was centered in the image. Lastly, we skull-stripped the scans using the HD-BET tool to remove any non-brain tissue [25].

## 2.3 Augmentation

GANs usually take a lot of data to train effectively [26, 27]. However, in this study, only two images of size 256x256x128 pixels were used. Therefore, using data augmentation was crucial. We used the batchgenerators framework for this task [28]. Since our model does not require coregistered images, this had to be accounted for in the data augmentation. Hence, we shifted and rotated the images in all three dimensions so the network learns the representation of the brain in space. Each training image was randomly rotated between -15° and 15° and shifted between 0 and 10 pixels in all three dimensions. Lastly, Gaussian noise was added to all images with the mean=0 and the variance ranging uniformly between 0 and 0.1.

## 2.4 Data

In this study two different datasets were used. The first was a private dataset including longitudinal follow-up scans from 15 patients diagnosed with recurrent Grade IV glioblastoma. As described in Kleesiek et al. [29], the baseline scan was defined as the scan before de novo treatment after tumor recurrence. The image resolution was 256x256x128 pixels. There were 13 male and 2 female patients with a mean age of 55.1 years. Image acquisition was performed on a 3 Tesla MRI scanner (Magnetom Verio, Siemens Healthcare, Erlangen, Germany).

The second was a publicly available dataset from the Cancer Imaging Archive (TCIA) [30], called Brain-Tumor-Progression [31]. This dataset includes two multi-channel MRIs each for 20 patients newly diagnosed with glioblastoma. The resolution of the images varied between 260x320x21 and 512x512x24 pixels. The model was developed on the first three patients, therefore only the last 17 patients were included in the evaluation. For both datasets only the T1-contrast-enhanced (T1ce) channels were used in this study.

## 2.5 Segmentation network for ground truth

To evaluate the proposed model's performance, ground truth segmentations were created. We used the neural network of the winner of the 2020 BRATS challenge for brain tumor segmentation for this task [32]. The segmentations contain three classes: enhancing tumor, edema and necrosis. Only the enhancing tumor class was used in this paper.

## 2.6 RANO classification

To further evaluate our model, we predicted a modified RANO classification. The RANO criteria for glioma is a radiological classification used to evaluate the treatment of glioblastoma [33]. We slightly modified this grading to allow for a classification using just the total enhancing tumor volume and disregarding any clinical information. The two classes, complete and partial response, were combined into one class called response. This class is defined as a reduction in tumor volume of more than 50%. Progression is defined as a growth in tumor volume of 25% or more. Consequently, stable disease is a change in tumor volume not corresponding to response or progression. The tumor volume was calculated in voxels.

The segmentations created by the BRATS network were again used to calculate the ground truth. Since the maps often showed a lot of noise at the edge of the brain, as shown in Figure 2, the outer 10 pixels in each dimension were disregarded. While this is potentially harmful for tumors at the edge of the brain, the advantages of removing the noisy regions outweigh the disadvantages. We created additional ternary maps from our network with just the three classes -1, 0 and 1. Voxels with a value smaller than -0.15 were defined as -1, showing tumor reduction and voxels with a value bigger than 0.15 were defined as 1, showing tumor growth. Classifications with a connected voxel count of 30 or less were set to 0 to remove some noise. The ternary map of each patient was added up to get the absolute change in tumor volume. This was added to the total tumor volume of the first timepoint to predict the volume of the second timepoint.

# 3 Results

## 3.1 Qualitative Assessment and Heatmaps

Figure 2 displays representative examples from both datasets. The map shows the changes in contrast-enhancing tumor in a reliable manner. The regions of tumor growth are represented as black (values <1 in the map). The regions of tumor reduction are represented as white (values >1 in the map). Converted to heatmaps they can be used to highlight the key regions of tumor growth/reduction.

As one can see, there are some recurring regions of noise in the maps. For example, the region next to the ventricular system is incorrectly noted as changed in either direction in most cases. Additionally, the edge of the brain often contains a lot of noise, as highlighted in Figure 2C. This can be a problem for tumors located at the edge of the brain or the ventricles.

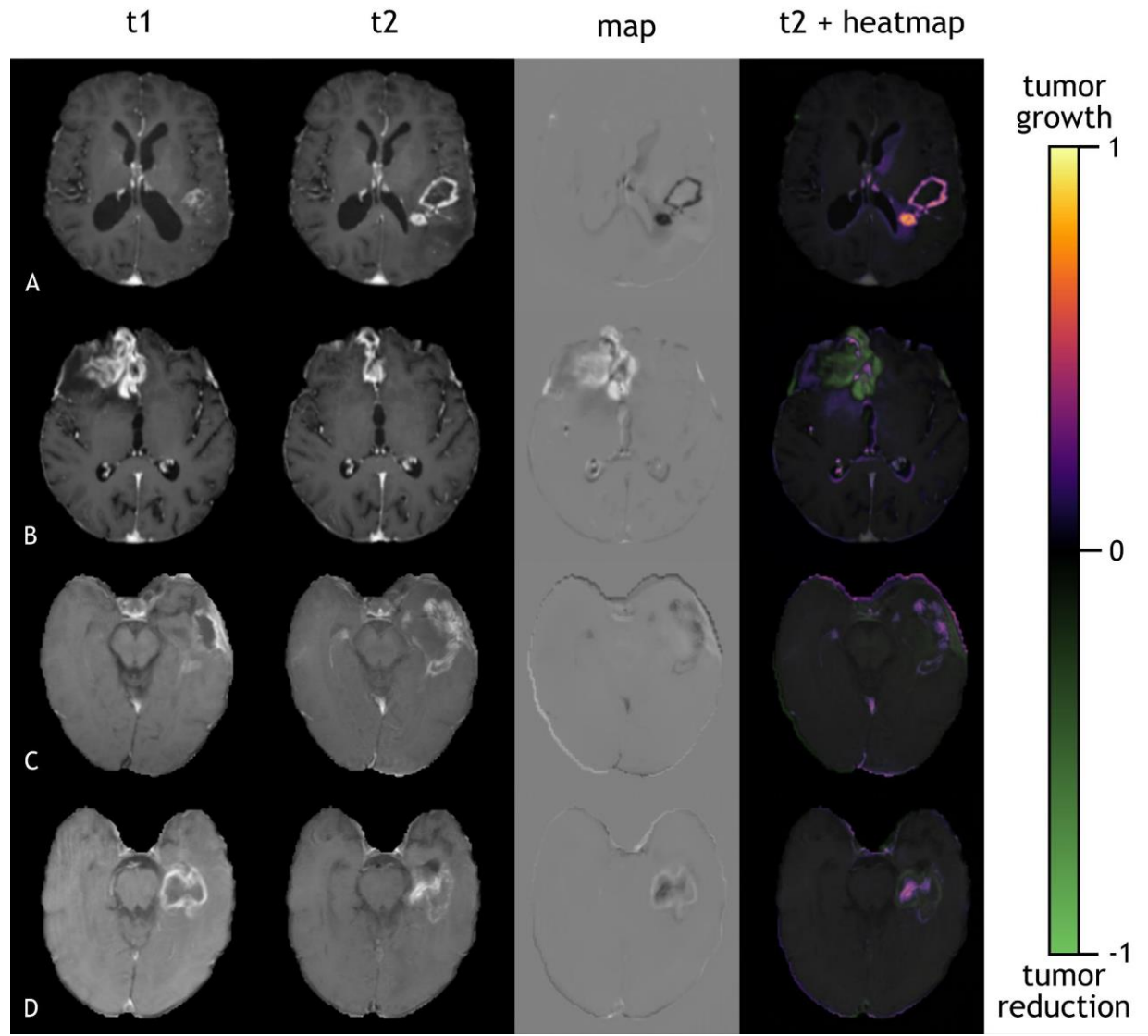

Figure 2: Examples of the T1ce images at different timepoints along with the calculated map. The last column shows heatmaps on top of the second time point to highlight key regions of change. A and B are from the private dataset, C and D are from the public dataset.

## 3.2 ROC analysis

An ROC analysis was performed to evaluate the model's prediction accuracy. The segmentations created by the BRATS network were used to calculate the ground truth. To get the classes tumor growth and reduction, the segmentation of the first time point was subtracted from the second time point.

The 2-class ROC analysis is shown in Figure 3. The area under the curve for tumor growth and reduction is 0.87 and 0.86, respectively. The micro-average AUC is 0.87. The AUC varied in the two datasets. In the private dataset the average AUC is 0.94. In the public dataset the average AUC is 0.76.

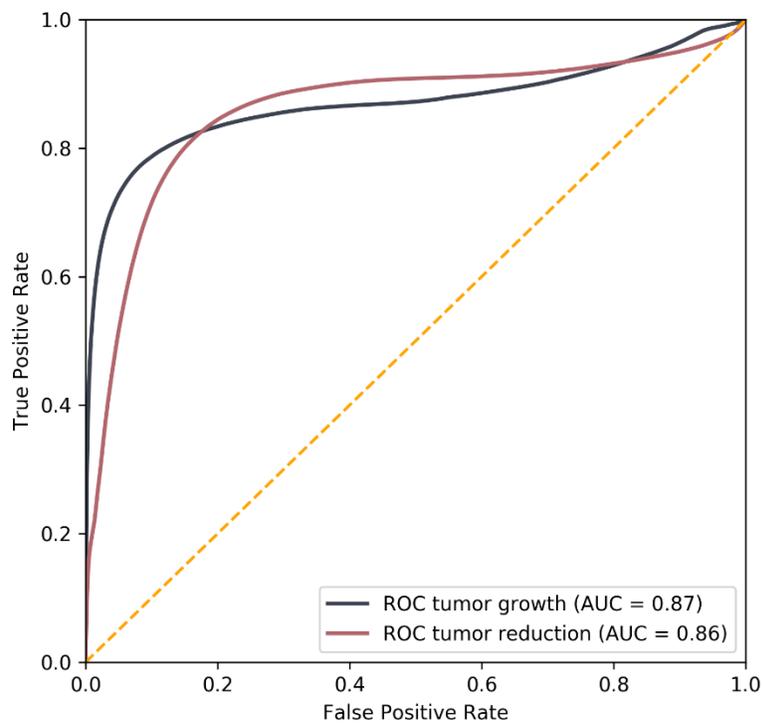

Figure 3: ROC Analysis for the prediction of tumor change compared to the ground truth of BRATS winning network nnUNet.

## 3.3 RANO classification

The results for the RANO classification are shown in Table 1. The total accuracy for the modified RANO classes was 65.6%. The performance for the two datasets was comparable with the total accuracy being 66.7% for the private dataset and 64.7% for the public dataset.

| RANO category | Accuracy |
|---|---:|
| Response | 70% |
| Stable disease | 80% |
| Progression | 50% |
| **Total** | **65.6%** |

Table 1: Accuracy of the prediction of modified RANO criteria for glioblastoma.

## 4 Discussion

In this contribution, we propose "*A net for everyone*", a personalized neural network that is trained with longitudinal data from a single patient. We designed and implemented a Wasserstein-GAN-based approach that works with only two scans from the same patient without any extra training data in an unsupervised fashion. That means, our method does not need any small or large quantities of datasets, and also does not need any manual or semi-manual annotations for training.

Alongside a qualitative evaluation, we show that the model achieves a high area under the curve in an ROC analysis, when compared to a state of the art deep learning model. It also shows that the model's performance for tumor growth and tumor reduction is very similar. The accuracy for the private dataset was significantly larger than for the public dataset. This can be explained by the difference in quality, as the public data was older and had a lower resolution, especially in the third dimension. Additionally, there were artifacts in some of the images, like parts of the brain were cut off. We implemented a modified RANO criteria, resulting in a combined accuracy of 66%. The generated heatmaps can aid in the diagnostic process to quickly find the key regions of interest.

It should be noted that the performance of deep learning models usually scales with the size of the dataset [34]. Therefore, this approach has an inherent disadvantage compared to classical supervised learning models with big datasets. However, using only the data of one patient comes with some advantages. First, our method is a privacy-safe approach. Medical records and medical image data are very sensitive and our approach stays within the same patient for the algorithmic training and execution. Second, getting large datasets in medical imaging has proven to be a challenging task due to these privacy concerns, and our method does not rely on this.

Furthermore, no registration is necessary for the training of our approach, which is a mandatory and crucial step in most approaches [35]. There are different methods for image registration, with some being completely automatic and others needing some manual input [36]. While these registration methods can be accurate for scenarios, like rigid registrations, especially deformable registrations are still challenging and there are problems with outliers [37]. These include post-surgery scans or patients with a different anatomy due to a large tumor. Both could lead to registration artifacts, which would compromise the further training. Our model does not need a separate registration step, avoiding these potential sources of errors.

The model does not explicitly learn to recognize changes in the tumor, but learns to recognize any changes between two images. However, since the contrast enhancing regions of the tumor are typically amongst the most intense regions in a T1ce scan, changes in these regions are particularly visible in the created maps, highlighting changes in tumor enhancement patterns. However, the proposed approach comes with two disadvantages that can be addressed in future research. First, any structural change in the brain not lying in the tumor will be recognized by the model. For example, a midline shift caused by tumor growth will cause changes in healthy regions of the brain and might be interpreted as growth or reduction of contrast enhancing tumor. This can also be interpreted as an advantage to point out all changes to the reader. Second, the model is prone to noise at the edge of the brain and next to the ventricles. The ventricles differ between two scans depending on the current cerebral spinal fluid volume. At the edge of the brain, the two scans also differ slightly due to the skull stripping. To account for the noise at the edge of the brain, we disregarded the outer pixels in the calculation of the modified RANO criteria. This is obviously a concern for tumors located in the cortex of the brain as it might cut out regions of the tumor. However, glioblastoma are typically located in the centrum semiovale, so in most cases this should not be a problem [38].

It should be noted that the ground truth from this work was not created by medical experts but by a neural network. However, the network used achieved a Dice Score for the enhancing tumor of 82% [32]. This lies within the range of the inter-rater variability of human raters of 74-85% [39], suggesting that medical experts would not change the ground truth significantly.

However, despite the above mentioned limitations, this study is a proof of concept that personalized neural networks can serve as a privacy-safe method to analyze longitudinal imaging data of a single patient in an unsupervised fashion. It has been shown that bidirectional brain tumor measurements that are used in the current RANO criteria are prone to underestimation of tumor growth on average, and even overestimation of growth for very small tumors [40, 41]. Therefore, having an efficient method for measuring the 3D tumor volume is essential for preoperative surgical planning, intraoperative neuro-navigation, and treatment monitoring [42, 43]. Lastly, the produced heatmaps can be a big help in the diagnosis of the MRI images, as they lead the reader directly to the key regions of changes.

Summarized, we proposed a deep learning architecture to create personalized neural networks. This study serves as is a proof of concept to show that training data from just one patient can be used to monitor tumor change in longitudinal MRI scans. Areas of future work include the application to other pathologies, such as aortic aneurysms and aortic dissections [44], where disease monitoring over several image acquisitions plays an important role.

## 5   Code availability

The source code will be uploaded to the following GitHub repository:

https://github.com/cstrack/personalized_nets_gbm

## 6   Supplemental Materials

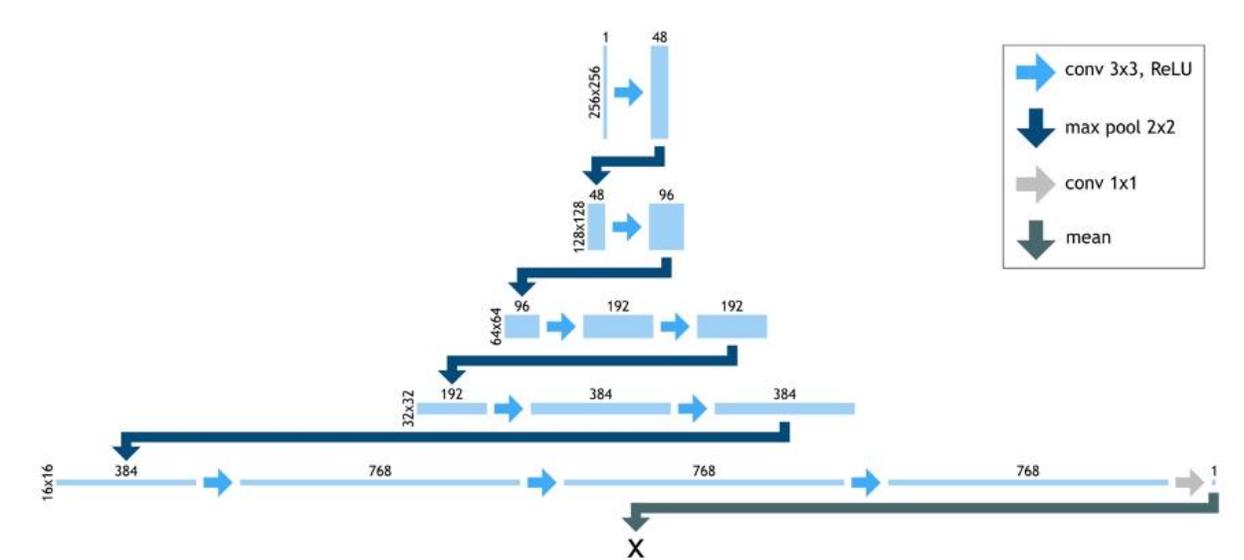

Figure 1: Architecture of the discriminator network. An encoder structure is used. The image size at each level is shown on the left of the blocks, the feature size is shown on top of the blocks. The number of 3x3 convolutions at each level increases further down the network. 2x2 max pool functions are used for downsizing. At the last level, the number of features is decreased to one with a 1x1 convolution. This last block is averaged to get the single value output of x.

**Acknowledgement**

We acknowledge the support of the REACT-EU project KITE (Plattform für KI-Translation Essen, EFRE-0801977, https://kite.ikim.nrw/).

## References

1. Deng, J.; Dong, W.; Socher, R.; Li, L.-J.; Li, K.; Fei-Fei, L. ImageNet: A Large-Scale Hierarchical Image Database. In Proceedings of the 2009 IEEE Conference on Computer Vision and Pattern Recognition; June 2009; pp. 248–255.
2. Fredrikson, M.; Jha, S.; Ristenpart, T. Model Inversion Attacks That Exploit Confidence Information and Basic Countermeasures. In Proceedings of the Proceedings of the 22nd ACM SIGSAC Conference on Computer and Communications Security; ACM: Denver Colorado USA, October 12 2015; pp. 1322–1333.
3. Vinyals, O.; Blundell, C.; Lillicrap, T.; Kavukcuoglu, K.; Wierstra, D. Matching Networks for One Shot Learning 2017.
4. Taigman, Y.; Yang, M.; Ranzato, M.; Wolf, L. DeepFace: Closing the Gap to Human-Level Performance in Face Verification. In Proceedings of the 2014 IEEE Conference on Computer Vision and Pattern Recognition; IEEE: Columbus, OH, USA, June 2014; pp. 1701–1708.
5. Zarrin, P.S.; Wenger, C. Implementation of Siamese-Based Few-Shot Learning Algorithms for the Distinction of COPD and Asthma Subjects. In Proceedings of the Artificial Neural Networks and Machine Learning – ICANN 2020; Farkaš, I., Masulli, P., Wermter, S., Eds.; Springer International Publishing: Cham, 2020; pp. 431–440.
6. Goodfellow, I.J.; Pouget-Abadie, J.; Mirza, M.; Xu, B.; Warde-Farley, D.; Ozair, S.; Courville, A.; Bengio, Y. Generative Adversarial Networks. *ArXiv14062661 Cs Stat* **2014**.
7. Kwon, G.; Han, C.; Kim, D. Generation of 3D Brain MRI Using Auto-Encoding Generative Adversarial Networks 2019.
8. Chuquicusma, M.J.M.; Hussein, S.; Burt, J.; Bagci, U. How to Fool Radiologists with Generative Adversarial Networks? A Visual Turing Test for Lung Cancer Diagnosis 2018.
9. Rubin, M.; Stein, O.; Turko, N.A.; Nygate, Y.; Roitshtain, D.; Karako, L.; Barnea, I.; Giryes, R.; Shaked, N.T. TOP-GAN: Stain-Free Cancer Cell Classification Using Deep Learning with a Small Training Set. *Med. Image Anal.* **2019**, *57*, 176–185, doi:10.1016/j.media.2019.06.014.
10. Lei, B.; Xia, Z.; Jiang, F.; Jiang, X.; Ge, Z.; Xu, Y.; Qin, J.; Chen, S.; Wang, T.; Wang, S. Skin Lesion Segmentation via Generative Adversarial Networks with Dual Discriminators. *Med. Image Anal.* **2020**, *64*, 101716, doi:10.1016/j.media.2020.101716.
11. Holland, E.C. Glioblastoma Multiforme: The Terminator. *Proc. Natl. Acad. Sci. U. S. A.* **2000**, *97*, 6242–6244, doi:10.1073/pnas.97.12.6242.
12. Harbeck, N.; Gnant, M. Breast Cancer. *Lancet Lond. Engl.* **2017**, *389*, 1134–1150, doi:10.1016/S0140-6736(16)31891-8.
13. Litwin, M.S.; Tan, H.-J. The Diagnosis and Treatment of Prostate Cancer: A Review. *JAMA* **2017**, *317*, 2532–2542, doi:10.1001/jama.2017.7248.
14. Adamson, C.; Kanu, O.O.; Mehta, A.I.; Di, C.; Lin, N.; Mattox, A.K.; Bigner, D.D. Glioblastoma Multiforme: A Review of Where We Have Been and Where We Are Going. *Expert Opin. Investig. Drugs* **2009**, *18*, 1061–1083, doi:10.1517/13543780903052764.
15. Egger, J.; Kapur, T.; Fedorov, A.; Pieper, S.; Miller, J.V.; Veeraraghavan, H.; Freisleben, B.; Golby, A.J.; Nimsky, C.; Kikinis, R. GBM Volumetry Using the 3D Slicer Medical Image Computing Platform. *Sci. Rep.* **2013**, *3*, 1364, doi:10.1038/srep01364.
16. Egger, J.; Pepe, A.; Gsaxner, C.; Jin, Y.; Li, J.; Kern, R. Deep Learning—a First Meta-Survey of Selected Reviews across Scientific Disciplines, Their Commonalities, Challenges and Research Impact. *PeerJ Comput. Sci.* **2021**, *7*, e773, doi:10.7717/peerj-cs.773.
17. Fu, Y.; Lei, Y.; Wang, T.; Curran, W.J.; Liu, T.; Yang, X. Deep Learning in Medical Image Registration: A Review. *Phys. Med. Biol.* **2020**, *65*, 20TR01, doi:10.1088/1361-6560/ab843e.

18. Arjovsky, M.; Chintala, S.; Bottou, L. Wasserstein GAN. *ArXiv170107875 Cs Stat* **2017**.
19. Baumgartner, C.F.; Koch, L.M.; Tezcan, K.C.; Ang, J.X.; Konukoglu, E. Visual Feature Attribution Using Wasserstein GANs. *ArXiv171108998 Cs* **2018**.
20. Ronneberger, O.; Fischer, P.; Brox, T. U-Net: Convolutional Networks for Biomedical Image Segmentation. *ArXiv150504597 Cs* **2015**.
21. Tran, D.; Bourdev, L.; Fergus, R.; Torresani, L.; Paluri, M. Learning Spatiotemporal Features with 3D Convolutional Networks. In Proceedings of the 2015 IEEE International Conference on Computer Vision (ICCV); December 2015; pp. 4489–4497.
22. Gulrajani, I.; Ahmed, F.; Arjovsky, M.; Dumoulin, V.; Courville, A. Improved Training of Wasserstein GANs. *ArXiv170400028 Cs Stat* **2017**.
23. Kingma, D.P.; Ba, J. Adam: A Method for Stochastic Optimization 2017.
24. Fedorov, A.; Beichel, R.; Kalpathy-Cramer, J.; Finet, J.; Fillion-Robin, J.-C.; Pujol, S.; Bauer, C.; Jennings, D.; Fennessy, F.; Sonka, M.; et al. 3D Slicer as an Image Computing Platform for the Quantitative Imaging Network. *Magn. Reson. Imaging* **2012**, *30*, 1323–1341, doi:10.1016/j.mri.2012.05.001.
25. Isensee, F.; Schell, M.; Pflueger, I.; Brugnara, G.; Bonekamp, D.; Neuberger, U.; Wick, A.; Schlemmer, H.-P.; Heiland, S.; Wick, W.; et al. Automated Brain Extraction of Multisequence MRI Using Artificial Neural Networks. *Hum. Brain Mapp.* **2019**, *40*, 4952–4964, doi:10.1002/hbm.24750.
26. Nuha, F.U.; Afiahayati Training Dataset Reduction on Generative Adversarial Network. *Procedia Comput. Sci.* **2018**, *144*, 133–139, doi:10.1016/j.procs.2018.10.513.
27. Ferreira, A.; Li, J.; Pomykala, K.L.; Kleesiek, J.; Alves, V.; Egger, J. GAN-Based Generation of Realistic 3D Data: A Systematic Review and Taxonomy 2022.
28. Isensee, F.; Jäger, P.; Wasserthal, J.; Zimmerer, D.; Petersen, J.; Kohl, S.; Schock, J.; Klein, A.; Roß, T.; Wirkert, S.; et al. Batchgenerators - a Python Framework for Data Augmentation 2020.
29. Kleesiek, J.; Petersen, J.; Döring, M.; Maier-Hein, K.; Köthe, U.; Wick, W.; Hamprecht, F.A.; Bendszus, M.; Biller, A. Virtual Raters for Reproducible and Objective Assessments in Radiology. *Sci. Rep.* **2016**, *6*, 25007, doi:10.1038/srep25007.
30. Clark, K.; Vendt, B.; Smith, K.; Freymann, J.; Kirby, J.; Koppel, P.; Moore, S.; Phillips, S.; Maffitt, D.; Pringle, M.; et al. The Cancer Imaging Archive (TCIA): Maintaining and Operating a Public Information Repository. *J. Digit. Imaging* **2013**, *26*, 1045–1057, doi:10.1007/s10278-013-9622-7.
31. Schmainda, K.; Prah, M. Data from Brain-Tumor-Progression 2019.
32. Isensee, F.; Jaeger, P.F.; Full, P.M.; Vollmuth, P.; Maier-Hein, K.H. *NnU-Net for Brain Tumor Segmentation*; arXiv, 2020;
33. Wen, P.Y.; Macdonald, D.R.; Reardon, D.A.; Cloughesy, T.F.; Sorensen, A.G.; Galanis, E.; DeGroot, J.; Wick, W.; Gilbert, M.R.; Lassman, A.B.; et al. Updated Response Assessment Criteria for High-Grade Gliomas: Response Assessment in Neuro-Oncology Working Group. *J. Clin. Oncol.* **2010**, *28*, 1963–1972, doi:10.1200/JCO.2009.26.3541.
34. Hestness, J.; Narang, S.; Ardalani, N.; Diamos, G.; Jun, H.; Kianinejad, H.; Patwary, M.M.A.; Yang, Y.; Zhou, Y. Deep Learning Scaling Is Predictable, Empirically 2017.
35. Erdt, M.; Steger, S.; Sakas, G. Regmentation: A New View of Image Segmentation and Registration. **2012**, 23.
36. Wyawahare, M.V.; Patil, D.P.M.; Abhyankar, H.K. Image Registration Techniques: An Overview. *Image Process. Pattern Recognit.* **2009**, *2*, 18.
37. Qin, B.; Gu, Z.; Sun, X.; Lv, Y. Registration of Images with Outliers Using Joint Saliency Map. *IEEE Signal Process. Lett.* **2010**, *17*, 91–94, doi:10.1109/LSP.2009.2033728.
38. Rees, J.H.; Smirniotopoulos, J.G.; Jones, R.V.; Wong, K. Glioblastoma Multiforme: Radiologic-Pathologic Correlation. *RadioGraphics* **1996**, *16*, 1413–1438, doi:10.1148/radiographics.16.6.8946545.

39. Menze, B.H.; Jakab, A.; Bauer, S.; Kalpathy-Cramer, J.; Farahani, K.; Kirby, J.; Burren, Y.; Porz, N.; Slotboom, J.; Wiest, R.; et al. The Multimodal Brain Tumor Image Segmentation Benchmark (BRATS). *IEEE Trans. Med. Imaging* **2015**, *34*, 1993–2024, doi:10.1109/TMI.2014.2377694.
40. Berntsen, E.M.; Stensjøen, A.L.; Langlo, M.S.; Simonsen, S.Q.; Christensen, P.; Moholdt, V.A.; Solheim, O. Volumetric Segmentation of Glioblastoma Progression Compared to Bidimensional Products and Clinical Radiological Reports. *Acta Neurochir. (Wien)* **2020**, *162*, 379–387, doi:10.1007/s00701-019-04110-0.
41. Dempsey, M.F.; Condon, B.R.; Hadley, D.M. Measurement of Tumor "Size" in Recurrent Malignant Glioma: 1D, 2D, or 3D? *AJNR Am. J. Neuroradiol.* **2005**, *26*, 770–776.
42. Fyllingen, E.H.; Stensjøen, A.L.; Berntsen, E.M.; Solheim, O.; Reinertsen, I. Glioblastoma Segmentation: Comparison of Three Different Software Packages. *PLOS ONE* **2016**, *11*, e0164891, doi:10.1371/journal.pone.0164891.
43. Sorensen, A.G.; Batchelor, T.T.; Wen, P.Y.; Zhang, W.-T.; Jain, R.K. Response Criteria for Glioma. *Nat. Clin. Pract. Oncol.* **2008**, *5*, 634–644, doi:10.1038/ncponc1204.
44. Pepe, A.; Li, J.; Rolf-Pissarczyk, M.; Gsaxner, C.; Chen, X.; Holzapfel, G.A.; Egger, J. Detection, Segmentation, Simulation and Visualization of Aortic Dissections: A Review. *Med. Image Anal.* **2020**, *65,* 101773, doi:10.1016/j.media.2020.101773.